\documentclass[conference]{IEEEtran}
\IEEEoverridecommandlockouts
\usepackage{amsmath,amssymb,amsfonts}

\usepackage[utf8]{inputenc}
\usepackage[english]{babel}
\usepackage[sorting=none]{biblatex}
\usepackage{subcaption}
\usepackage{csquotes}
\usepackage{algorithmic}
\usepackage{graphicx}
\usepackage{graphics}

\usepackage{textcomp}
\usepackage{xcolor}
\def\BibTeX{{\rm B\kern-.05em{\sc i\kern-.025em b}\kern-.08em
    T\kern-.1667em\lower.7ex\hbox{E}\kern-.125emX}}
\graphicspath{ {./Figs/} }

\begin{document}

\title{On exploiting the synaptic interaction properties to obtain frequency-specific neurons}

\author{\IEEEauthorblockN{Guillaume Marthe}
\IEEEauthorblockA{\textit{CITI, EA3720} \\
\textit{Univ Lyon, INSA Lyon, Inria}\\
69621 Villeurbanne, \\ France \\
guillaume.marthe@insa-lyon.fr}
\and
\IEEEauthorblockN{Claire Goursaud}
\IEEEauthorblockA{\textit{CITI, EA3720} \\
\textit{Univ Lyon, INSA Lyon, Inria}\\
69621 Villeurbanne, \\ France \\
claire.goursaud@insa-lyon.fr}
\and
\IEEEauthorblockN{Romain Cazé}
\IEEEauthorblockA{\textit{IEMN} \\
\textit{UMR 8520}\\
59650 Villeneuve-d'Ascq, \\France \\
romain.caze@univ-lille.fr}
\and
\IEEEauthorblockN{Laurent Clavier}
\IEEEauthorblockA{\textit{CNRS, UMR 8520 - IEMN} \\
\textit{IMT Nord Europe,} \\ \textit{Université de Lille}\\
59000 Lille, France \\
laurent.clavier@imt-nord-europe.fr}
}

\IEEEoverridecommandlockouts
\IEEEpubid{\makebox[\columnwidth]{979-8-3503-9918-9/23/\$31.00 ©2023 IEEE \hfill}
\hspace{\columnsep}\makebox[\columnwidth]{ }}

\maketitle

\IEEEpubidadjcol

\begin{abstract}

Energy consumption remains the main limiting factors in many IoT applications.
In particular, micro-controllers consume far too much power. In order to overcome this problem, new circuit designs have been proposed and the use of spiking neurons and analog computing has emerged as it allows a very significant consumption reduction. 
However, working in the analog domain brings difficulty to handle the sequential processing of incoming signals as is needed in many use cases.
In this paper, we use a bio-inspired phenomenon called Interacting Synapses to produce a time filter, without using non-biological techniques such as synaptic delays. We propose a model of neuron and synapses that fire for a specific range of delays between two incoming spikes, but do not react when this Inter-Spike Timing is not in that range. We study the parameters of the model to understand how to choose them and adapt the Inter-Spike Timing. The originality of the paper is to propose a new way, in the analog domain, to deal with temporal sequences.  

\end{abstract}


\section{Introduction}

The Internet of Things (IoT) brings a lot of promises for everyday life applications \cite{InternetOT}. 
One significant limiting issue remains the energy consumption of the devices: lifespans of several years or tens of years cannot be achieved as soon as a minimum of intelligence has to be deported into the devices. 
However, as most devices involved in the IoT only send data sporadically. Thus, to reduce this consumption a key solution is to let them sleep as much as possible and only wake them up when they need to interact. Nonetheless, it is still important to be able to address the devices (whether it is the network, i.e. a gateway, or a neighbouring device) at any time without depending on a global synchronisation of the system.

This can be achieved through the use of wake-up receivers (WuR), circuits whose role is to continuously monitor the channel and wake up the main receiver only when a specific signal is received \cite{WUR}. WuR leads to a decrease of the energy wasted between communications. 

An essential condition for their use is that the listening consumption is extremely low, but many challenges remain unresolved \cite{WuR2022}.
Up to now, those WuR still rely on micro-controllers that can perform pattern recognition but consume power around $200 \mu W$ \cite{perfWUR}.
This is too high to allow the battery of the WuR to last the intended lifespan of the node 
\cite{lifetime}.

Meanwhile, Spiking Neural Networks (SNN) is a technology that has emerged in the field of signal processing in recent years as a low power consumption solution \cite{SNNvision}. They are mainly used for image recognition \cite{SNNvideo} or to study model brain activity in neuroscience \cite{insect}. This bio-inspired systems are based on transistors to mimic neuron behavior. They can compute as fast as actual devices but consume less \cite{4fJ}.

To reduce a step further the energy consumption, we consider in this paper specific fully analog circuits that reproduce the neurons' behavior 
\cite{sub35pW,Mart2303:Wake}. Until now, these neurons have been used for combinatorial approaches that do not involve time sequences. 
They consume even less energy than traditional SNNs but are more limited on applications because they restrict the usage we can make of the neuron. For example, it is highly complex to use synaptic delay without a digital circuit. The sequential applications such as vision, video recognition or signal processing were up to now impossible to realise. In this paper, we show that it is possible to introduce this time dependency. For this, we propose to exploit the biological phenomenon of synaptic interaction and its memory effect \cite{articleRomain}. The phenomenon we use corresponds in biology to the effect of dendrites on the cortical pyramidal neurons \cite{dendrites}. We exploit this phenomenon to make the neuron reacting to a specific time pattern and evaluate the neuron parameters on the selectivity. 

The paper is organized as follows: section \ref{section2} presents the neurons and system models and how they are simulated. 
Section \ref{section3} describes the neurons behavior and their implementation. 
The choice of the parameters and their impact are then studied in Section \ref{section4}. Finally, Section \ref{section5} concludes the paper.

\begin{figure*}[!t]
\centering
\includegraphics[scale=0.5]{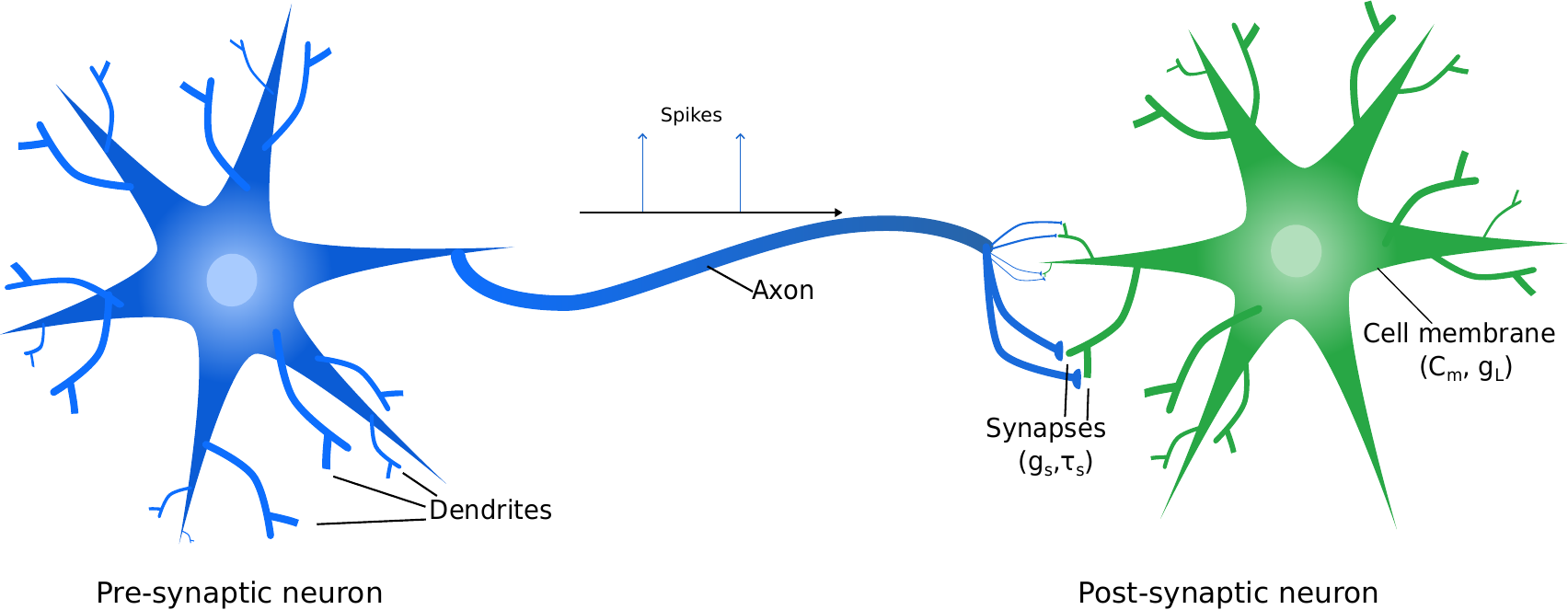}
\caption{Diagram of a biological neuron}
\label{fig:neuron}
\end{figure*}

\section{Models}\label{section2}

\subsection{LIF neuron with classical synapses}

A biological neuron 
receives spikes from the incoming synapses, 
aggregate them within its membrane voltage, and then 
sends forward a spike if the voltage has reached a given 
threshold. An illustration 
of neurons and synapse can be found in Fig. \ref{fig:neuron}.

A bio-inspired neuron is based on the same principle. It is modeled by its reaction according to the incoming electric spike train. The internal voltage varies depending on the incoming current. 
There exists several models to depict the relationship between the incoming current and the voltage. The Leaky Integrate and Fire (LIF) model is the reference, widely considered in the literature \cite{LIFmodel}, it is expressed by :
\begin{equation}\label{eq2 : LIF}
    C_m \frac{dv}{dt} = - g_L(v(t) - v_{rest}) + I(t)
\end{equation}
with $C_m$ the membrane capacity, $v(t)$ the membrane potential at time $t$, $v_{rest} = -65 mV$ the resting membrane potential and $g_L$ the membrane conductance. 
$I(t)$ models the time-varying synaptic input current, which can be written in case this incoming signal is a sequence of spikes as: 
\begin{equation}\label{eq3 : spikeTrain}
   I(t) = \sum_{i} \delta (t - t_i)
\end{equation}
with $t_i$ corresponding to the different spike times, and $\delta$ the Dirac function.

Each time the voltage reaches the threshold $v_{th}$ a spike is triggered and the voltage is reset to $v_{rest}$. 
The gray curve in Fig. 2a represents the temporal evolution of the voltage over time when receiving 
a single spike. In a LIF, we observe an instantaneous growth at the reception of the spike, then an exponential decrease due to the leakage in the neuron.

\subsection{Saturating LIF}

The LIF model is a useful, yet simplified model. Indeed, there is an additional phenomenon that is usually neglected, but that we will exploit here.
In this model, called Saturating LIF (SLIF), the synapses are now saturating synapses. This models the fact that the synapse parameters are temporary modified after receiving a spike. 
In this case, the current no longer depends only on the current input but also on what has happened before. 
Thus, $I(t)$ \eqref{eq3 : spikeTrain} is now given by $I_s(t)$ such that:
\begin{equation}\label{eq3 : Is}
    I_s(t) = g_s(t) (E_s-v(t)) 
\end{equation}
The synaptic current depends on the difference between $v(t)$ the neuron voltage and $E_s = 0$ mV the synaptic reversal potential. $g_s(t)$ is the time-varying synapse conductance depending on the previous incoming spike pattern. It is bounded between 0 and the saturation value that we set at 100 pS for our simulations. Each time the synapse receives an incoming spike, $g_s(t)$ will rise to its maximal value, and decays exponentially following :

\begin{equation}\label{eq4 : dgs}
    \frac{dg_s(t)}{dt} = \frac{-g_s(t)}{\tau_s} 
\end{equation}

with $\tau_s$ the synaptic time constant.

This is called "Saturating Synapses" because we make $g_s(t)$ saturate to a specific value.

The consequence of this saturation is that the dynamic of the tension is constrained and makes it slower to reach the target membrane voltage as we can see on the grey curve in Fig. \ref{fig:4casesSSLIF}b.



We can observe on this curve in Fig. \ref{fig:4casesSSLIF} that the voltage increases smoothly, and then decreases until it reaches its resting potential. 
We are going to show how this can be used to create a short term memory for temporal computations.

\section{Proposition}\label{section3}

Our objective 
is to use this neuron structure to spike only at the reception of a specific firing rate among any incoming spike train. This allows to spike only for a given range of inter-spike duration or, said differently, to filter a specific period of spikes. 
To do so, we want to reach the highest voltage if two incoming spikes arrive within a given Inter-Spike Timing (IST) range. 

To clarify our proposal, we propose to study in more details the behavior of a neuron with this saturating synapse. 
To evaluate the phenomenon we send two consecutive spikes to the two types of neurons LIF and SLIF spaced by different IST which represent the time between the reception of those two spikes.

\begin{figure}[!t]
\centering
\includegraphics[scale=0.4]{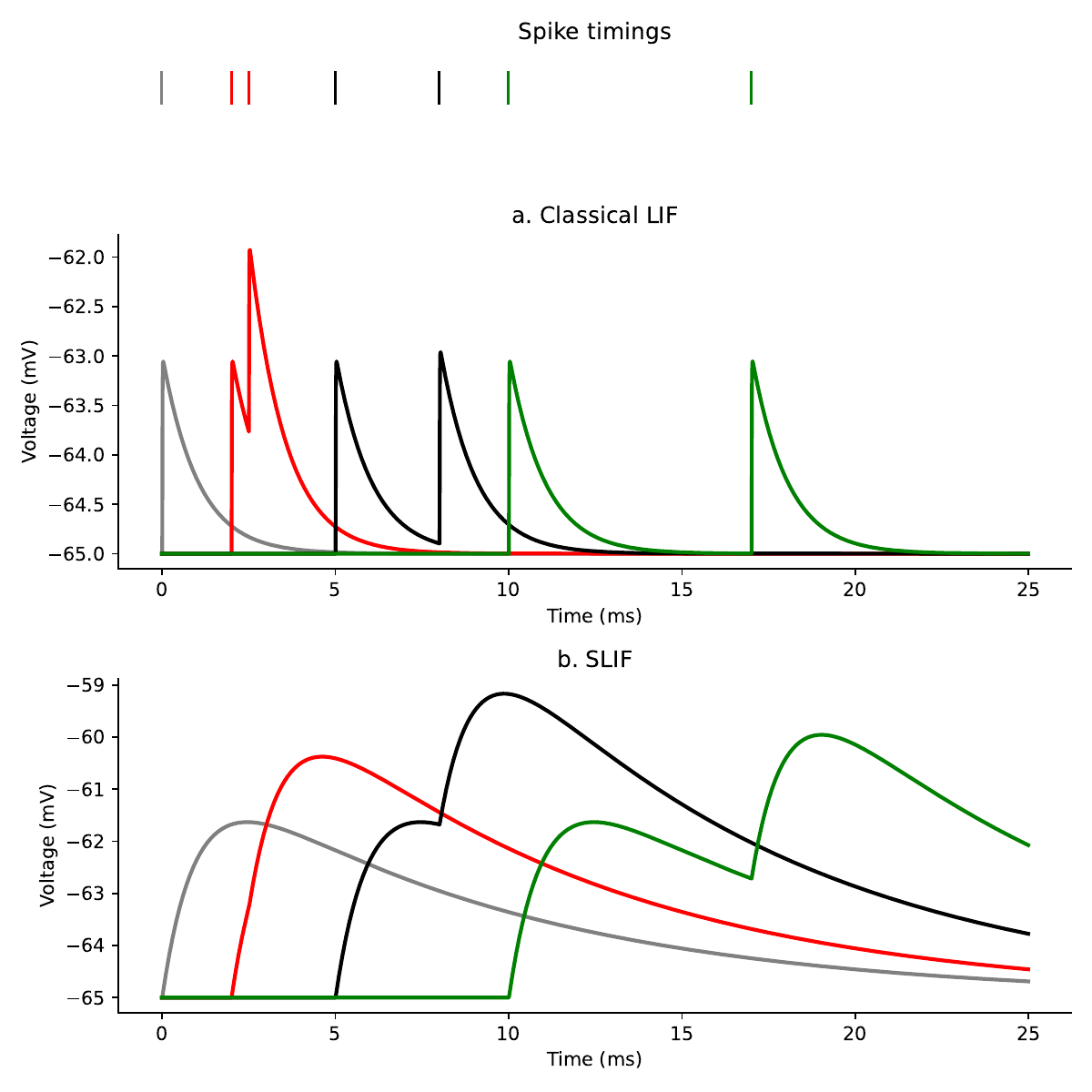}
\caption{Response of the neuron to different ISTs}
\label{fig:4casesSSLIF}
\end{figure}

The results are shown in Fig. \ref{fig:4casesSSLIF} with 4 curves each, corresponding to different configurations. The leftmost (grey) one was obtained by sending only 1 spike to the neuron. The 3 other curves were obtained by sending 2 spikes to the neuron, with a different IST. The 3 ISTs are from the left to the right: 0.5 ms, 3 ms and 7ms. The corresponding spike positions are reported on the top as reference.

When a classical LIF neuron receives a spike, its membrane potential suddenly rises. Then, its potential decays due to the leaky aspect of the neuron. When a second spike is sent, the potential rises again instantly of the same amplitude value. The highest amplitude is reached when the IST is minimal. Indeed, the potential leak is lower when it has less time to decay. On the opposite, if the second spike is sent a long time after the first one, the potential has decayed more, and then the amplitude is lower than in the first case. Thus, the amplitude decays when the IST increases. 


The response of the SLIF is presented on Fig. \ref{fig:4casesSSLIF}b in the same conditions. 
With the saturating synapses, if a second spike is received right after the first one, the synapse is still saturated and this spike does not make the potential rise of the same quantity the first one did.  
We can observe that, contrarily to the LIF case, the increase in the amplitude of the membrane voltage after the reception of second spike is not constant 
and depends on the IST. 
For the second curve from the left, the black one, the spikes are too close and the synaptic saturation limits the impact of the second spike, which will have only a small impact on the membrane potential. For the rightmost one (green), the spikes are too far from each other and the leaky aspect of the neuron 
makes the potential drop significantly 
before receiving the second spike. This will reduce the highest reached amplitude of the membrane voltage.

\begin{figure}[!t]
\centering
\includegraphics[scale=0.5]{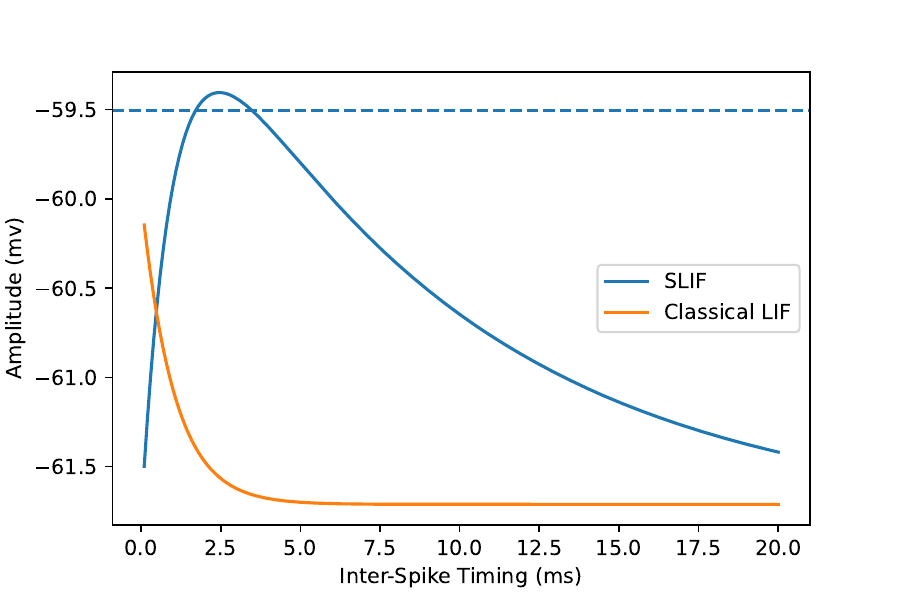}
\caption{Highest amplitude reached by a SLIF and a classical LIF neurons for different ISTs where $C_m = 10^{-5} \mu F.cm^{-2}$, $g_L = 5 \cdot 10^{-5} S.cm^{-2}$ and $\tau_s = 0.9 ms$}
\label{fig:ampIST}
\end{figure}

In addition, we have plotted the amplitudes reached as a function of IST in Fig. \ref{fig:ampIST}. With classical synapses we have had an exponentially decreasing and monotonous function. 
For the SLIF, we can see that the maximal amplitude is reached for an IST of 3ms and that there are two phases. For an IST lower than 3ms, the saturation leads to a lower amplitude than the optimum, with a linear rise until the favorite IST, and then a decrease due to the leaky part of the neuron. 

To further illustrates the impact of the saturating synapse, let us represent the value of the spiking threshold by the 
horizontal dashed line. 
We clearly understand that the neuron will spike only for a given range of IST so that the threshold can be used to discriminate a fine range of IST values. In this simulation, the neuron was spiking if the input spikes were spaced by an IST between 2 and 3.5 ms. The width of this range of ISTs will be called Timewidth (TW).
Finally, we can note that the IST of 3 ms leads to the highest increase. 

\begin{figure}[!t]
\centering
\includegraphics[scale=0.5]{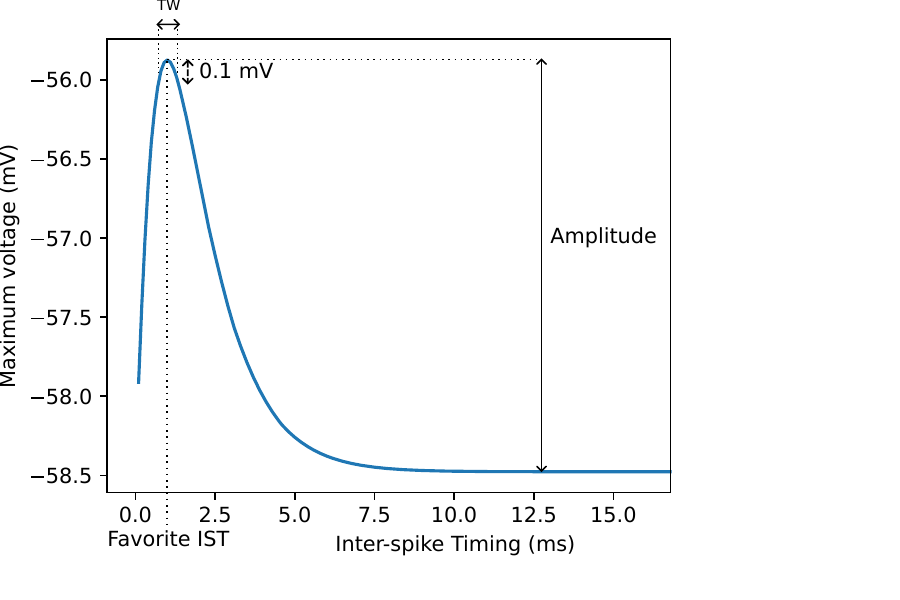}
\caption{Maximal amplitude reached by a SLIF where $C_m = 10^{-3} \mu F.cm^{-2}$, $g_L = 10^{-5} S.cm^{-2}$ and $\tau_s = 0.01 ms$}
\label{fig:usrefAmp}
\end{figure}

\begin{figure*}[!t]
\centering
\begin{subfigure}{0.45\textwidth}
    \includegraphics[width=\textwidth]{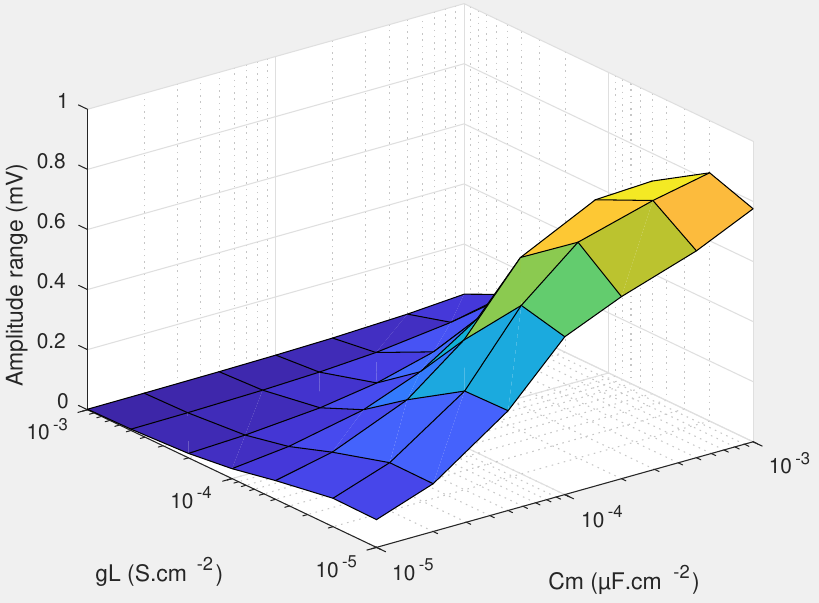}
    \caption{Maximum amplitude for the favorite IST and $\tau_s = 0.1 ms$}
    \label{fig:cmgLAmp}
\end{subfigure}
\hfill
\begin{subfigure}{0.45\textwidth}
    \includegraphics[width=\textwidth]{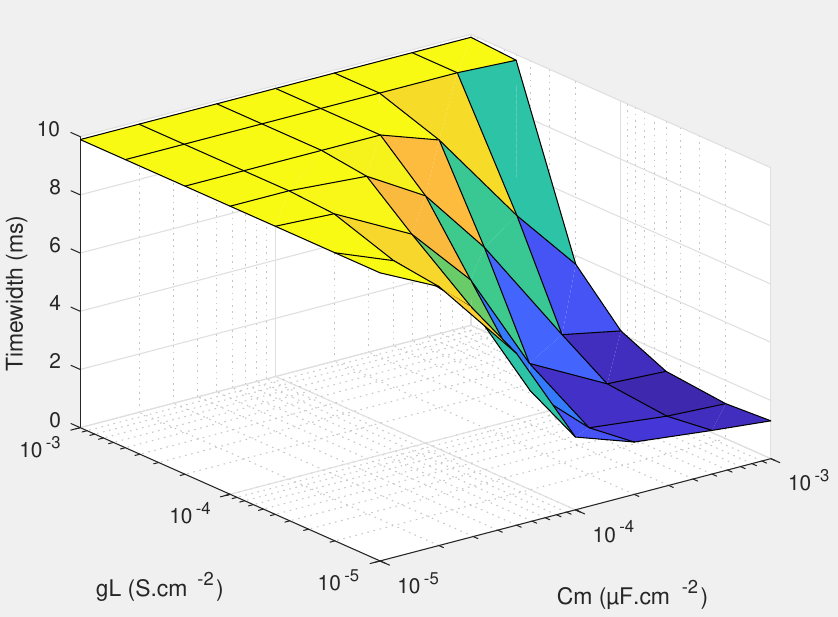}
    \caption{Timewidth around the favorite IST and $\tau_s = 0.1 ms$}
    \label{fig:cmgLTemp}
\end{subfigure}
\vfill
\begin{subfigure}{0.45\textwidth}
    \includegraphics[width=\textwidth]{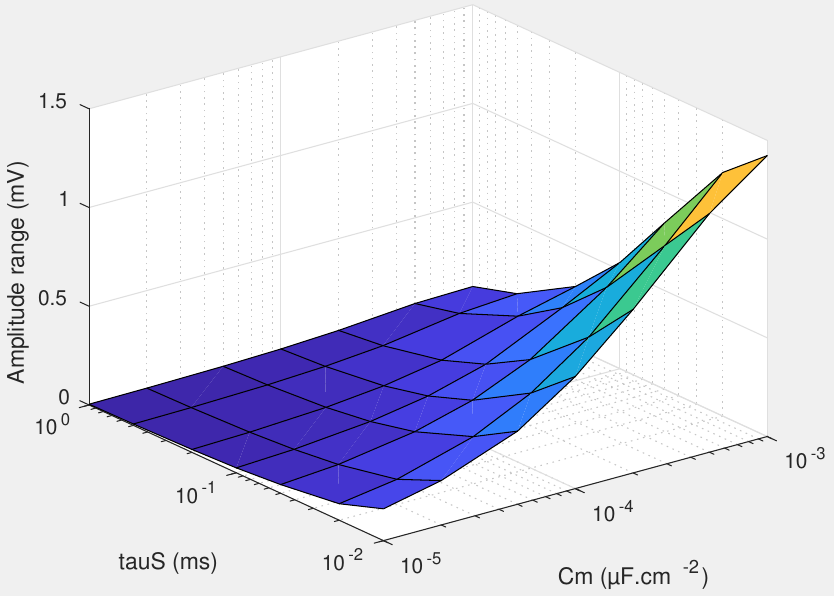}
    \caption{Maximum amplitude for the favorite IST and $g_L = 10^{-4} S.cm^{-2}$}
    \label{fig:cmtauSAmp}
\end{subfigure}
\hfill
\begin{subfigure}{0.45\textwidth}
    \includegraphics[width=\textwidth]{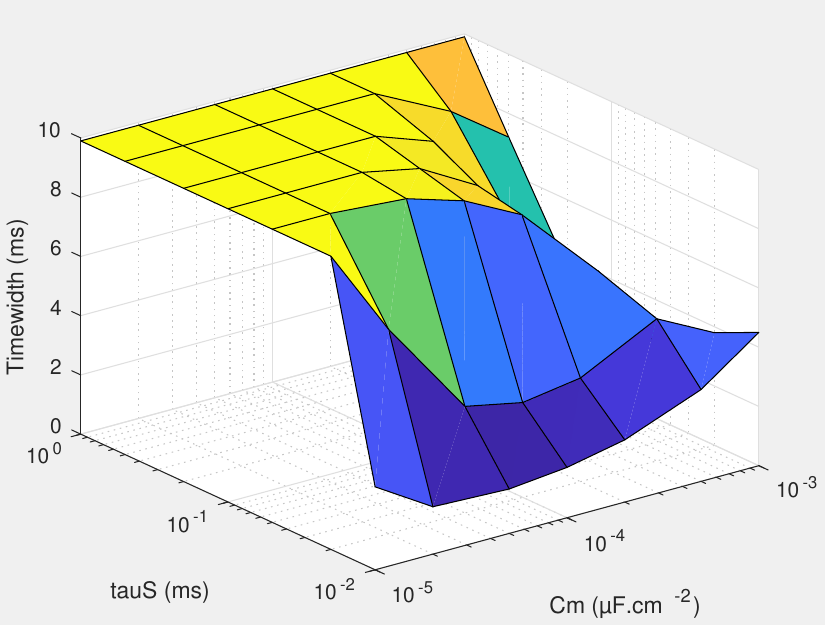}
    \caption{Timewidth around the favorite IST and $g_L = 10^{-4} S.cm^{-2}$}
    \label{fig:cmtauSTemp}
\end{subfigure}
\vfill
\begin{subfigure}{0.45\textwidth}
    \includegraphics[width=\textwidth]{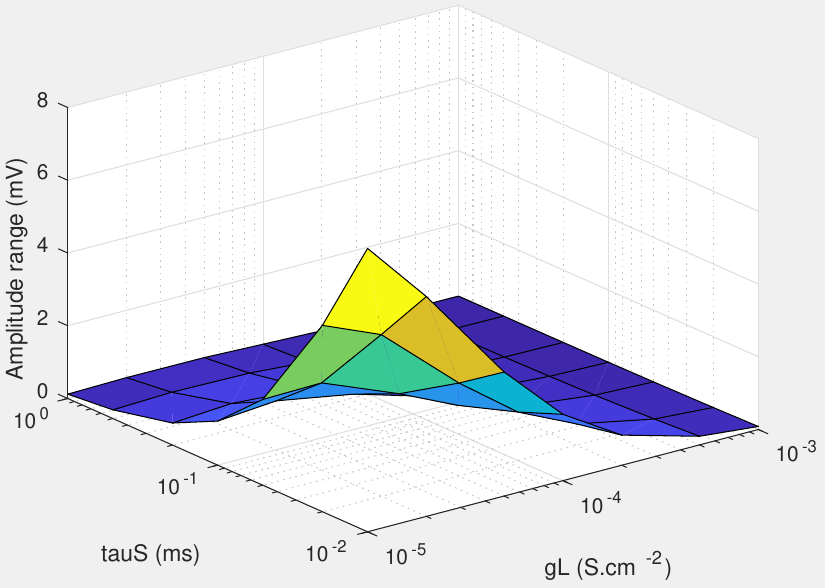}
    \caption{Maximum amplitude for the favorite IST and $C_m = 10^{-4} \mu F.cm^{-2}$}
    \label{fig:gLtauSAmp}
\end{subfigure}
\hfill
\begin{subfigure}{0.45\textwidth}
    \includegraphics[width=\textwidth]{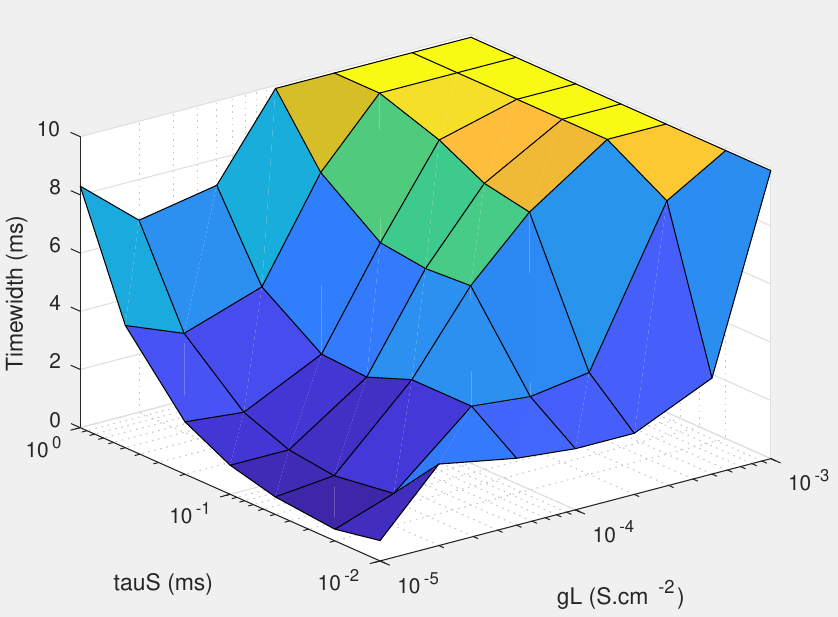}
    \caption{Timewidth around the favorite IST and $C_m = 10^{-4} \mu F.cm^{-2}$}
    \label{fig:gLtauSTemp}
\end{subfigure}
\caption{Evolution of Amplitude and TW according to $C_m$, $g_L$ and $\tau_s$}
\label{fig:figures}
\end{figure*}

The shape of the curve in Fig. \ref{fig:ampIST} (increase time, decrease time, maximum amplitude and when it occurs) 
are influenced by the parameters of the equations:
\newline

\begin{itemize}
    \item 
$C_m$, the membrane capacity which permits to scale the whole dynamic of the neuron. 
    \item 
$g_L$, the conductance of the neuron which regulates the leaky phenomenon for the membrane voltage. The bigger $g_L$ is, the fastest the voltage will return to its resting value $v_{rest}$.
    \item 
$g_S$, the synaptic conductance. Its value is impacted by the parameter $\tau_S$ and modifies the speed of evolution of the membrane potential, when a spike is received. Therefore, $\tau_S$ is the parameter which permits to choose how fast $g_S$ will return to its default value, here $0$. In our model, we consider that $g_S$ value is clipped between $0$ and $100pS$.
\end{itemize}

By modifying the previous parameters, we can affect the neuron response. For example, in Fig. \ref{fig:usrefAmp}, the new parameter set permits to switch from a millisecond IST range to a microsecond range and thinned the curve, while keeping the amplitude constant. 

We can see in Fig. \ref{fig:usrefAmp} the 3 main parameters describing the neuron behavior: 
the amplitude, the position of the TW and its width. The amplitude corresponds to the difference between the maximum and minimum voltage reached for the considered ISTs. To measure the TW, we set a threshold 0.1 mV under the maximal amplitude, and we look at the range of IST that makes the membrane voltage exceeds this threshold.

The objective is now to optimize the parameter set. The goal is to obtain the highest amplitude range (so as to be more robust to perturbations), while reducing TW (so as to be more precise in the selected rates).
The favorite IST to target depends on the frequency pattern chosen.

\section{Parameters study}\label{section4}

Our simulations were realised using the Python library Brian2 \cite{Brian2}.
We have seen that the SLIF is a model that permits to spike only for a specific range of IST values, depending of our parameters. Thus, we study here the impact of the parameters on both the amplitude and TW. 
To realize those studies, we have set the parameters at a reference set which is $C_m = 10^{-4} \mu F.cm^{-2}$, $g_L = 10^{-4} S.cm^{-2}$ and $\tau_s = 0.1 ms$. We then varied two parameters simultaneously between 0.1 to 10 times its original value.

As a reminder, we try to optimise the thinness of the TW and the maximal amplitude to distinguish better the responses to different ISTs.

\subsection{Membrane capacity}

As it directly impacts the dynamic of the neuron, we first analyze $C_m$ to determine the order of magnitude of the favorite ISTs and then the other parameters to adjust the amplitude value and the TW. $C_m$ is the membrane capacity and it is multiplied to the dynamics of the neuron potential $\frac{dv}{dt}$. We can see in Fig. \ref{fig:cmgLAmp} and \ref{fig:cmtauSAmp} that when $C_m$ increases, the maximal amplitude of the voltage rises.
 
For TW, we can see in Fig. \ref{fig:cmgLTemp} and \ref{fig:cmtauSTemp} that if $C_m$ increases, TW is reduced. Thus increasing $C_m$ permits to have a more precise identification of the spike rate. Finally, Fig. \ref{fig:CmgLIST} and \ref{fig:CmtauSIST} show that the position of the favorite IST evolves in the same direction as $C_m$. When it increases, the IST take larger values.

\subsection{Membrane conductance}

\begin{figure}[!t]
\begin{subfigure}{0.45\textwidth}
    \includegraphics[width=\textwidth]{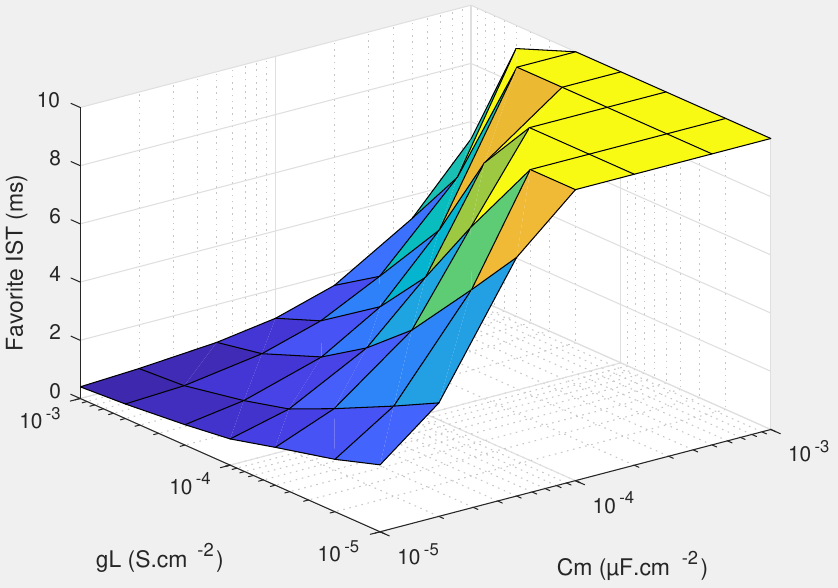}
\caption{Favorite IST with $\tau_s = 0.1 ms$}
\label{fig:CmgLIST}
\end{subfigure}
\vfill
\begin{subfigure}{0.45\textwidth}
    \includegraphics[width=\textwidth]{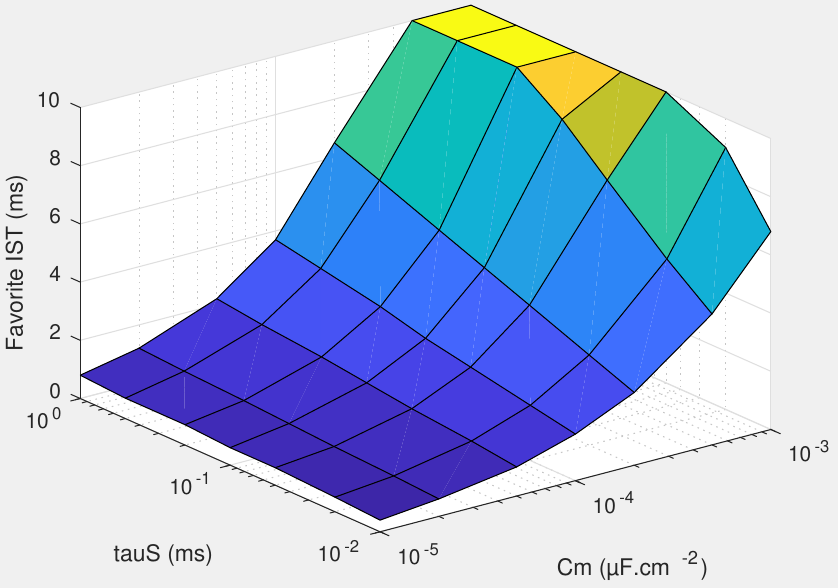}
\caption{Favorite IST with $g_L = 10^{-4} S.cm^{-2}$}
\label{fig:CmtauSIST}
\end{subfigure}
\vfill
\begin{subfigure}{0.45\textwidth}
    \includegraphics[width=\textwidth]{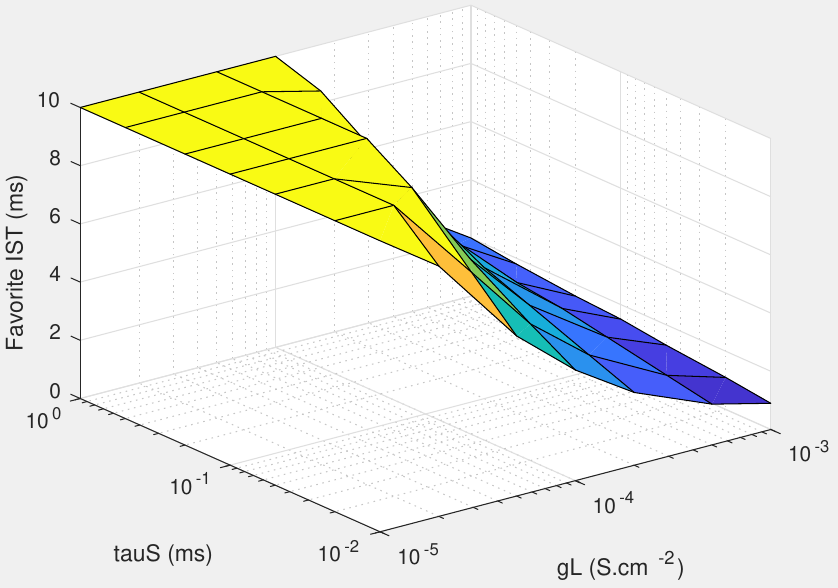}
    \caption{Favorite IST with $C_m = 10^{-4} \mu F.cm^{-2}$}
    \label{fig:gLtauSIST}
\end{subfigure}
\caption{Evolution of Amplitude, IST and TW according to $g_L$ and $\tau_s$}
\label{fig:figures1}
\end{figure}

$g_L$ is the membrane conductance which defines the dynamics of the membrane potential leak. It is directly proportional to how fast the membrane voltage comes back to $v_{rest}$ after receiving a spike. We can see in Fig. \ref{fig:cmgLAmp} and \ref{fig:gLtauSAmp} that when $g_L$ increases, the maximum amplitude decreases as intended. Indeed, the more $g_L$ rises, the more the leak phenomenon is strong, and so, the sooner we come back to the resting potential. Thus, the IST is reduced and the TW is thinner, as we can see respectively in Fig. \ref{fig:CmgLIST} and \ref{fig:gLtauSIST}, and Fig. \ref{fig:cmgLTemp} and \ref{fig:gLtauSTemp}.




\subsection{Time constant}

The value of $g_L$ is used to scale the maximal amplitude we want to reach. We can then modify $\tau_S$ to adjust the TW more precisely.

$\tau_S$ is the time constant of $g_s$, which is the synaptic conductance. After receiving a spike, $g_s$ instantly rises until it reaches its upper bound. It will then decay to its lower bound. The speed of the decay is ruled by $\tau_S$. 
$\tau_S$ is impacting $g_s$ which can be associated as a weight in the model. Thus, when $\tau_s$ has a low value, $g_s$ varies faster and then get bigger variations. This enhances the effect of the second spike on the membrane potential, because $g_s$ saturates less than for a higher value of $\tau_S$ as shown in Fig. \ref{fig:cmtauSAmp} and \ref{fig:gLtauSAmp}.
As it rules the membrane voltage gain from the second spike, $\tau_S$ is influencing TW. The higher $\tau_S$ gets, the wider the TW becomes. Indeed, $g_s$ varies less so the amplitude of potential decays slower. Thus, we will use $\tau_S$ to adapt the range of IST making the neuron spike. 

However, we can observe in Fig. \ref{fig:CmtauSIST} and \ref{fig:gLtauSIST} that the evolution of $\tau_s$ has almost no impact of the favorite IST. That is another reason to scale it at the end, to only change the TW when the IST is set.

Finally, one may note that interestingly, there is no need to compromise between amplitude range and TW, improving one also improves the other.

\section{Conclusion}\label{section5}

In this paper, we studied a new bio-inspired spiking neuron model based on saturating synapses called SLIF. We showed that our neuron responds differently based on the Inter-Spike Timing of the incoming spikes. Unlike the LIF neuron with classical synapses, the SLIF allows us to reach a maximal membrane voltage amplitude for an IST that can be configured. We have exploited this natural phenomenon to transform a spiking neuron and its synapses into a temporal filter that emits a spike only for a chosen IST.
Then, we studied the impact of the main parameters of our model on the amplitude of the membrane voltage and the timewidth of the IST that can represent the selectivity of the model.
Future works will analyze how SLIF can allow to discriminate spike trains based on the delay between each spikes to develop an ultra low-power wake-up radio. 

\section*{Acknowledgment} This work was supported by the research project U-WAKE, financed by the French National Research Agency (ANR).

\printbibliography[
heading=bibintoc,
title={Bibliography}] 

\end{document}